\documentclass[preprints,article,accept,moreauthors,pdftex]{Definitions/mdpi}

\usepackage{siunitx}
\usepackage{array}
\usepackage{multirow}
\usepackage{subfigure}

\makeatletter
\renewcommand{\@thesubfigure}{\normalsize(\textbf{\alph{subfigure}})}
\makeatother

\usepackage[linesnumbered,ruled,vlined]{algorithm2e}
\setitemize{parsep=6pt,itemsep=0pt,leftmargin=*,labelsep=5.5mm}
\setenumerate{parsep=6pt,itemsep=0pt,leftmargin=*,labelsep=5.5mm}
\setlist[description]{itemsep=0mm}
\SetAlgoCaptionSeparator{ }

\SetKwInput{KwInput}{Input}                
\SetKwInput{KwOutput}{Output}              
\SetKwInput{Kwp}{Procedure}
\firstpage{1}
\pubvolume{9}
\issuenum{10}
\articlenumber{1700}
\pubyear{2020}
\copyrightyear{2020}
\history{Received: 11 September 2020; Accepted: 12 October 2020; Published: 16 October 2020}
\updates{yes} 

\Title{ Sky Imager-Based Forecast of Solar Irradiance Using Machine Learning }

\Author{Anas Al-lahham 
$^{1,}$*\href{https://orcid.org/0000-0001-8413-6957}{\orcidicon}, Obaidah Theeb $^{1}$\orcidB{}, Khaled Elalem $^{1}$\orcidC{}, Tariq A. Alshawi $^{1}$ and Saleh~A.~Alshebeili~$^{1,2}$}

\AuthorNames{Anas Al-lahham, Obaidah Theeb, Khaled Elalem, Tariq  Alshawi and Saleh Alshebeili}

\address{%
$^{1}$ \quad Electrical Engineering Department, King Saud University, Riyadh 11421, Saudi Arabia; entheeb@outlook.com (O.T.); Khaledalem71@gmail.com (K.E.); talshawi@ksu.edu.sa (T.A.A.); dsaleh@ksu.edu.sa (S.A.A.)\\
$^{2}$ \quad King Abdulaziz City for Science and Technology (KACST)-Technology Innovation Center (TIC) in Radio Frequency and Photonics (RFTONICS), King Saud University, \mbox{Riyadh 11421, Saudi Arabia}}

\corres{Correspondence: anas.hkj@outlook.com}

\abstract{Ahead-of-time forecasting of the output power of power plants is essential for the stability of the electricity grid and ensuring uninterrupted service. However, forecasting renewable energy sources is difficult due to the chaotic behavior of natural energy sources. This paper presents a new  approach to  estimate short-term solar irradiance from sky images. The~proposed algorithm extracts features from sky images and use learning-based techniques to estimate the solar irradiance. The~performance of proposed machine learning (ML) algorithm is evaluated using two publicly available datasets of sky images. The~datasets contain over 350,000 images for an interval of 16 years, from 2004 to 2020, with the corresponding global horizontal irradiance (GHI) of each image as the ground truth. Compared to the state-of-the-art computationally heavy algorithms proposed in the literature, our approach achieves competitive results with much less computational complexity for both nowcasting and forecasting up to 4 h ahead of time.}


\begin{document}


\section{Introduction}
Photovoltaic (PV) systems have attained a rapid increase in popularity and utilization to face the challenges of climate change and energy insecurity, as~they bring a potential displacement for fossil fuel due to its merits of being pollution-free and its role of limiting global warming. However, the~volatility and uncertainty of solar power resources are some of the main challenges that affect the PV power output, which, along with inaccurate forecasting, may impact the stability of the power grid~\cite{marcos2011power,martinez2016value}.
Therefore, accurate irradiance forecasting may help power system operators to perform different actions in the grid operation, such as load following, scheduling of spinning reserves or unit commitment~\cite{sediqi2019stochastic}. \par PV power output mainly depends on the amount of solar irradiance on a collection plane. However, the~amount of solar irradiance is affected by various weather conditions such as clouds and dust. Thus, solar irradiance may be prone to rapid fluctuations in various regions~\cite{kleissl2013solar}.
Various models have been proposed to forecast solar irradiance; these forecasting models are classified into parametric and statistical models. The~main difference between these two models is the dependency on historical data; the parametric, physical or ``white box'' models do not need any historical data to generate the prediction of solar irradiance. They generate the prediction according to meteorological processes and weather conditions, such as cloud formation, wind, and~temperature. The~most well-known physical model is the numerical weather prediction (NWP), which, as~the time horizon increases, offers greater accuracy over statistical models. Hybrid methods are also popular as they combine a mix of both models~\cite{antonanzas2016review}.
\par Several physical and statistical methods have been proposed in the literature for solar irradiance forecasting. Larson et al. \cite{larson2016day} proposed a methodology to generate a day-ahead power output forecast of two PV plants using publicly available NWP from two models; a PV physical model was allowed to obtain power output using global horizontal irradiance (GHI) values obtained from the two models.
The statistical and machine learning (ML) models predict solar irradiance by extracting relations among historical data to train the model; therefore, a~decent training sample is essential in order to produce an accurate model. There are two well known statistical methods, artificial intelligence (AI) techniques and regressive methods, which are mostly used for short term forecasting (less than 4 h). In~such cases, NWP does not perform well because of the lack of the necessary granularity as a way to add future information to forecasting models. \par Talha A. Siddiqui et al. \cite{siddiqui2019deep} presented a deep neural network approach to forecast short-term solar irradiance. The~datasets in that work were collected from two different locations. The~first location was the  Solar Radiation Research Laboratory (SRRL) (Golden, Colorado dataset), where an image was recorded using a Total-Sky Imager commercial camera (TSI) every 10 min with a mechanical sun tracker to prevent satiety in the image. The~dataset was collected from 2004--2016, and~the total images captured totaled 304,309. The~second location was in Tucson, Arizona, 
where the dataset had been recorded at the Multiple Mirror Telescope Observatory (MMTO). The~dataset spans the period from the months of November 2015 to May 2016. That paper applied two types of irradiance predictions on the datasets, namely nowcasting and forecasting.  The~forecasting was for a duration up to 4~h. Air~temperature, wind speed, relative humidity and other auxiliary data were used to improve the quality of the model. The~work in~\cite{siddiqui2019deep} used the normalized mean absolute percentage error (nMAPE)  metric to quantify the prediction accuracy. The~proposed algorithm uses computationally heavy ML~techniques.\par Anto Ryu et al. \cite{ryu2019preliminary} presented an approach for short-term solar irradiance forecasting for 5--20~min ahead, using (TSI), with~two forecasting models. First, a~Convolutional Neural Network (CNN) model was used with only sky images taken by TSI. Second, a~CNN model using both sky images and lagged GHI are used as input data. Moreover, the~output of estimated GHI of the first model was used as input data to the second model. A~third persistence model was used to compare the forecasting accuracy of the proposed CNN models.
\par Graeme Vanderstar et al. \cite{vanderstar2018solar} proposed a method to forecast two hours ahead of solar irradiance using Artificial Neural Network (ANN). The~use of different remote solar monitoring stations is combined with the use of ML concepts using genetic algorithms. The~algorithm was used to find the best selection of solar monitoring stations chosen from the 20 available sites. The~algorithm has a forecasting capability using a small number of monitoring stations---as few as five~stations.

\par Ke-Hung Lee et al. \cite{lee2018solar} presented a method for short-term solar irradiance forecasting using electromagnetism-like neural networks. The~results of the electromagnetism-like neural network were compared with the backpropagation neural network. The~comparison results showed that the prediction of the electromagnetism-like neural network was better than the backpropagation neural~network.
\par M.Z. Hassan et al. \cite{hassan2017forecasting} conducted research into the forecasting of day-ahead solar radiation using ML approach. That paper collected the datasets samples from local solar power plant at Nadi Airport in Fiji. The~average values of the solar power in a known duration were contained in the data samples. The~authors of~\cite{hassan2017forecasting} implemented two regression techniques, one was the linear least squares and the second was the Support Vector Machine (SVM). Multiple kernel functions with SVM were used to obtain good results on non-linear separable data. Mean Absolute Error (MAE) and Root Mean Square Error (RMSE) were considered as prediction accuracy metrics. The~results showed that no forecasting algorithm of the proposed models can be perfect for all~conditions.

\par An extensive literature review on the prediction of PV power production was conducted by Ahmed et al. \cite{ahmed2020review}. This comprehensive review included different forecasting methods, input correlation analysis, uncertainty quantification, time stamp, data pre- and post-processing, forecast horizon, network optimization, performance evaluations, weather classification and extensive reviews of ANN and other AI 
techniques. This review shows that the conventional and statistical forecasting methods in terms of reliability, accuracy and computational economy could not outrun the ML approach in the form of ANN or its hybrids, especially for short-term forecast horizons. The~complexity and computational time of ML models were also considered in this study, showing that having multiple inputs significantly increases complexity and computational~time.

\par  Huynh et al. \cite{huynh2020near} developed a model to forecast global solar radiation (GSR). The~model is based on deep learning principle---more accurately, the long short-term memory (LSTM) network modelling strategy. This model considers  very short-term forecasting (1--30 min forecasting horizon). This study claims that the LSTM model outperforms other deep learning models, a~statistical model, a~single hidden layer, and~a machine learning-based model. Hybridization with other models is also considered in this study to further improve the performance of the LSTM~model.


As reported in the literature, the~forecasting accuracy of solar irradiance remains less adequate. State-of-the-art deep learning solar irradiance prediction algorithms have demonstrated excellent performance but are computationally heavy. On~the other hand, a~major shortcoming of parametric models is the high dependency on NWP, which  is  spatially too coarse to accurately predict solar irradiance due to the generality of the information provided by weather forecast as well as the lack sufficient spatial and temporal resolution~\cite{antonanzas2016review}.\par In this paper, we develop new computationally efficient ML algorithms for forecasting the solar irradiance for durations from 1 h up to 4 h. This study targets accurate prediction of  GHI by training multiple forecasting models, using sky images obtained from the  SRRL. The~GHI ground truth for the sky images are obtained from a measurement and instrumentation data center (MIDC) in Golden, Colorado ~\cite{Anu:2013,nerl}. The~main contributions of this paper are as follows:

\begin{itemize}[leftmargin=*,labelsep=5.8mm]
\item	Proposing a prediction approach that does not rely on  meteorological parameters, and~encodes an input sky image to take the form of a one-dimensional (1-D) vector  to facilitate the use of less complex ML regressors.
\item	Adopting Latent Semantic Analysis (LSA) to
reduce the size of the regressor input vector, without~decreasing
the prediction accuracy.
\item	Evaluating the performance of a new proposed approach using a 350,000-sample dataset. The~results show that the proposed approach outperforms the more complex state-of-the-art forecasting methodology presented in~\cite{siddiqui2019deep}.
\end{itemize}

The development of algorithms that are computationally efficient and solely rely on sky images for irradiance prediction will enable their implementation in inexpensive off-the-shelf hardware platforms. The~organization of this paper is as follows. The~background about the data collection is given in Section~\ref{sec2}. Section~\ref{sec3} presents the proposed GHI prediction algorithms. The~results and discussion are given in Section~\ref{sec4}. Our concluding remarks are outlined in Section~\ref{sec5}.


\section{Data~Collection}
\label{sec2}

Sky images are obtained from a wide-angle lensed Sky Imager. Measured GHI is taken from the MIDC. This dataset is used in this work to forecast GHI up to 4 h ahead of time. The~proposed algorithms are developed using two publicly available datasets of sky images captured in Golden, Colorado (39.742\si{\degree} N, 105.18\si{\degree} W, Colorado, USA). Golden, located in  north--central, Colorado, U.S., lies at an elevation of 1829 m, and is surrounded by mountains. It has a warm climate with a significant amount of rainfall during the year. The~datasets were recorded at SRRL~\cite{Anu:2013,nerl}. Samples of the obtained images are illustrated in Figure~\ref{Skyimage}. The~description for each of the datasets is as follows:


\begin{figure}[H]
\centering

\subfigure[~Sunny]{
\includegraphics[scale=0.6]{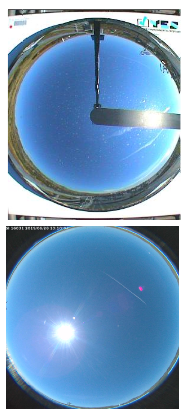}

}
\subfigure[~Cloudy]{
\includegraphics[scale=0.6]{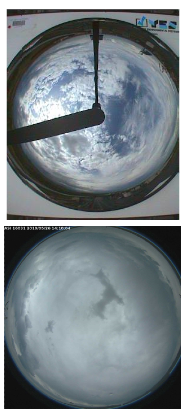}

}
\subfigure[~Rainy]{
\includegraphics[scale=0.6]{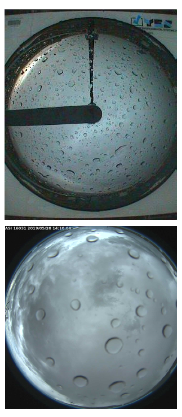}

}

\caption{Sky images for three types of weather conditions 
from the TSI-880 (\textbf{top}) and ASI-16~(\textbf{bottom})~datasets. }
\label{Skyimage}
\end{figure}
\unskip


\subsection{Total Sky Imager (TSI-880)}
The sky images in this dataset have been taken using Total Sky Imager model TSI-880. It provides full-color wide angle view sky images at an interval of 10 min. The~imager has a mechanical sun blocking band which tracks and blocks the sun, thus preventing saturation in the image.  It has been capturing all sky images since 14 July 2004~\cite{morris2005total}. We used 313,562 images for an interval of 12~years during the period from 14 July 2004 to 31 December 2016. The~first 261,092 images, which~were captured over 10 years, from~14 July 2004 to 31 December 2014, were assigned as a training set. The~remaining 52,470, covering 2 years from 2015 to 2016, were assigned as a testing set to evaluate the performance of both nowcasting and forecasting models. Other selections of training and testing sets were also~considered.

\subsection{All Sky Imager (ASI-16)}
This dataset was captured by All Sky Imager (ASI-16), an automatic full-sky camera system with a fisheye lens  for a 180\si{\degree} field of view. It provides full hemispheric pictures of the sky and clouds at an interval of 10 min. It has been capturing all sky images since 26 September 2017~\cite{ASI}. We used a total of 57,863 images from 26 September  2017 to 20 February 2020, such that 70\% of the total images were for training and 30\% were for testing. The~corresponding GHI data was obtained from the National Renewable Energy Laboratory (NREL) Baseline Measurement System (BMS), which has measured and logged data every minute since 15 July 1981. In~this paper, we consider 10 min GHI data obtained from pyranometer model CMP22~\cite{Anu:2013,nerl}.


\section{GHI Prediction~Algorithms }
\label{sec3}
The development of prediction algorithms is comprised of two main stages: feature extraction and regression. In~the first stage, the~features are extracted directly from the raw sky images  or their transformed versions.  In~the second stage, supervised learning is used to obtain the regression~models.

\subsection{Feature~Extraction}
Pre-processing steps are used to make the data suitable for ML algorithms. First, by~using the open-source software Python and OpenCV, a~sky image was converted into a three-dimensional matrix with dimensions 288 $\times$ 352 $\times$ 3. Second,  to~reduce the computational load, the~sky image was  downsized into an RGB 3-dimensional array of size \(M\times M\times 3\), where \(M\) = 32. Then, each image was reshaped into the form of a one-dimensional (1-D) array of size \(M^{2}\times 3\) samples (pixels). For~\(N\) available images, the~dataset's dimension becomes \(N\times (M^{2}\times 3)\) samples.

The input array feature vectors to the ML algorithms are pixel values. The~size of this vector is quite large as \(M\) = 32 in our development. In~this paper, LSA is used to reduce the number of features. LSA, also known as latent semantic indexing (LSI), introduced by Deerwester et al. \cite{deerwester1990indexing},  performs a  linear dimensionality reduction using the method of truncated singular value decomposition (SVD). Given a  rectangular matrix \(X\) of size \(Q\times D\),
the SVD of \(X\) is:
\begin{equation}
X=U \Sigma V^T
\end{equation}
where \(U\in \mathbb{C}^{Q\times Q}\) and \(V\in \mathbb{C}^{D\times D}\) are orthogonal
matrices, the~columns of \(U\) are called the left singular vectors of \(X\), while the columns of \(V\) are the right singular values of \(X\). \(\Sigma \in \mathbb{R}^{Q\times D}\) is the matrix containing the singular values of \(X\) along its diagonal. However, the~truncated SVD produces a low-rank approximation of \(X\) with the \(k\) largest singular values:
\begin{equation}
X\approx X_k=U_k \Sigma_k V^T_k
\end{equation}
where \(k<r\) (the number of non-zero singular values), \(U_k\in \mathbb{C}^{Q\times k}\) and \(V_k\in \mathbb{C}^{D\times k}\), \(\Sigma_k \in \mathbb{R}^{k\times k}\) \cite{mirzal2013limitation,cherkassky2007appendix,klema1980singular}. If~\(X\) is the training set with \(Q=L\) and \(D=M^{2}\times 3\times m\), then the reduced dimension training set will~be
\begin{equation}
X^\prime=U_k \Sigma^T_k
\end{equation}

This new transformed set contains \(L\times k\) features. In~the testing phase, the~input features matrix, \(T\), of~size \(l\times (M^{2}\times 3)\) pixels, is first transformed to a reduced form using
\begin{equation}
T^\prime= T V_k
\end{equation}

The transformed vector \(T^\prime\) is now of dimension \(l\times k\).

\subsection{Regression~Algorithms}
\label{sec3.2}
Two machine learning algorithms have been used to develop the regression models. In~particular, the~Random Forest (RF) and K-nearest neighbors (KNN) algorithms are considered. The~choice of these two algorithms has been based on extensive investigations to determine a regression algorithm with competitive performance and reduced computational~complexity.

\subsubsection{KNN}
KNN is one of the simplest of all ML algorithms, which can be used for both regression and classification. KNN finds the closest neighbors for a set of testing points based on a user-defined number called (\(K\)) within the given features. The~neighbors are picked from a set of training points whose classifications are noted. The~parameter \(K\) defines the number of nearest neighbors used for the regression. KNN could be considered as a lazy learning and non-parametric algorithm.  Choosing the value of \(K\) is essential to avoid the risk of overfitting. Without~tuning this parameter, we run the risk of having two noisy data points that are close enough to each other to outvote the right data points. The~values of K are fine tuned to be 2 using extensive experimentation with the implementation of cross validation techniques (CV) \cite{pal2020data}, and~using Euclidean distance as the distance function~\cite{shalev-shwartz_ben-david_2019,pedro2012assessment}.

\subsubsection{Random~Forest}
RF is considered one of the most famous ensemble ML techniques. It is used for performing both regression and classification tasks. It operates by constructing multiple numbers of decision trees. After~training, the~prediction for the test sample is done by averaging the predictions of all the decision trees. The~unique feature of the RF algorithm, which makes it different from other bagging algorithms, is that it selects random subsets of features at each split. This is beneficial because if one or more features are powerful in predicting the output target value, these features will be selected in building many of the next trees. The~RF algorithm avoids overfitting the decision trees on their training set using the bagging technique. Bagging selects random subsets of the training set to fit each tree. This procedure leads to a better performance since it decreases the variance of the model, without~increasing the bias~\cite{russell_2016,james2013introduction}. By~using extensive experimentations along with the CV techniques, the~different parameters of the RF algorithm are fine tuned to be as follows (No.trees = 200, maximum depth = 100, No.features (\(p\)) = $\sqrt{k}$), where \(p\) is the number of features to consider when looking for the best~split.

\subsection{Predictors'~Architectures}
\begin{figure}[H]
\captionsetup{justification=centering}
\centering
\includegraphics[width=15.6cm]{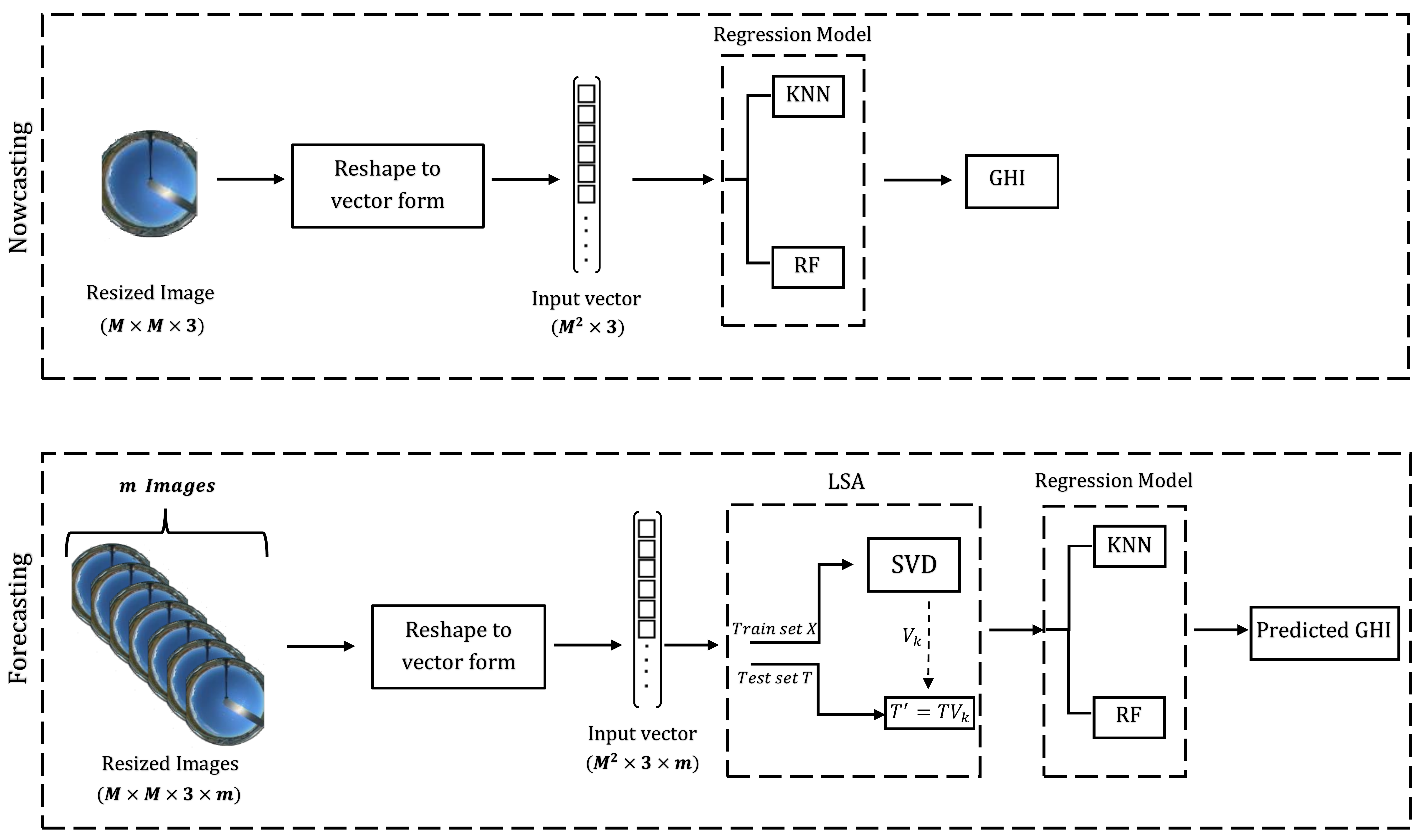}
\caption{The proposed  architectures for nowcasting and~forecasting. }
\label{block}
\end{figure}

The proposed architectures for both nowcasting and forecasting prediction operations are  illustrated in Figure~\ref{block}.

Nowcasting is the prediction of the solar irradiance at the instant the frame is captured~\cite{xingjian2015convolutional}. Each raw image in the dataset was down-sized  into an RGB three-dimensional array. This array is directly reshaped  into a 1-D input vector, which is applied to a regression model to predict solar irradiance~(GHI).

In forecasting, the~current image and \(m-1\) previous (look-back) images are used to form a concatenated input vector. Because~the resulting input vector large in size, LSA is used to reduce its dimensionality. Therefore, \(k\) features are extracted from the input vector and applied to the regression model. Algorithm \ref{algorithm1} shows the pseudocode of the forecasting process.

\vspace{12pt}
\begin{algorithm}[H]
\DontPrintSemicolon
\SetAlgoNoLine
\KwInput{Training set $X$ of size \(L \times (M^{2}\times 3)\).\\ Test set $T$ of size \(l \times (M^{2}\times 3)\).\\
\(N\)= \(L+l\)\\ Ground truth set G of size \(1\times L\)}

\KwOutput{Predicted GHI up to 4 h ahead}
\Kwp{}

Form look-back \(\hat{X}\) of size \((L-m+1) \times(M^{2}\times 3\times m)\)\\
Form look-back \(\hat{T}\) of size \((l-m+1) \times(M^{2}\times 3\times m)\)\\
Compute truncated  SVD : \(\hat{X}_k=U_k \Sigma_k V^T_k\)\\
Form transformed training train set \(X^\prime= U_k \Sigma^T_k\)\\
Form transformed  test set \(T^\prime= T V_k\)\\
Fit ML model over training set \(X^\prime\)\\
Generate forecast over test set \(T^\prime\)\\
Calculate nMAP error
\caption{Forecasting~Process }
\label{algorithm1}
\end{algorithm}
\vspace{12pt}


\par Three statistical metrics are used to assess the performance of the models using two different datasets. These metrics are as~follows.

The normalized mean absolute percentage error (nMAPE)
\begin{equation}
\text{nMAPE}=\sum_{i=1}^{l} \frac{|y_{i}-\hat{y}_{i}|}{\sum_{i=1}^{l} y_{i}}\times100
\end{equation}

The root mean square error (RMSE)
\begin{equation}
\text{RMSE}=\sqrt{\frac{1}{l}\sum_{i=1}^{l} ({y_{i}-\hat{y}_{i}})^2}
\end{equation}

The normalized root mean square error (nRMSE)
\begin{equation}
\text{nRMSE}=\frac{RMSE}{({y_{i_{max}}}-{y_{i_{min}}})}
\end{equation} where \(l\) is the number of testing samples, and~ \(y_{i}\) and \(\hat{y}_{i}\) are the true and predicted values of GHI, respectively.


\section{Results And~Discussion}
\label{sec4}
This section reports the results of the proposed prediction algorithms. Specifically, we consider  two types of predictions, nowcasting and~forecasting.

\begin{figure}[H]
\centering

\subfigure[~TSI-880]{
\includegraphics[scale=0.6]{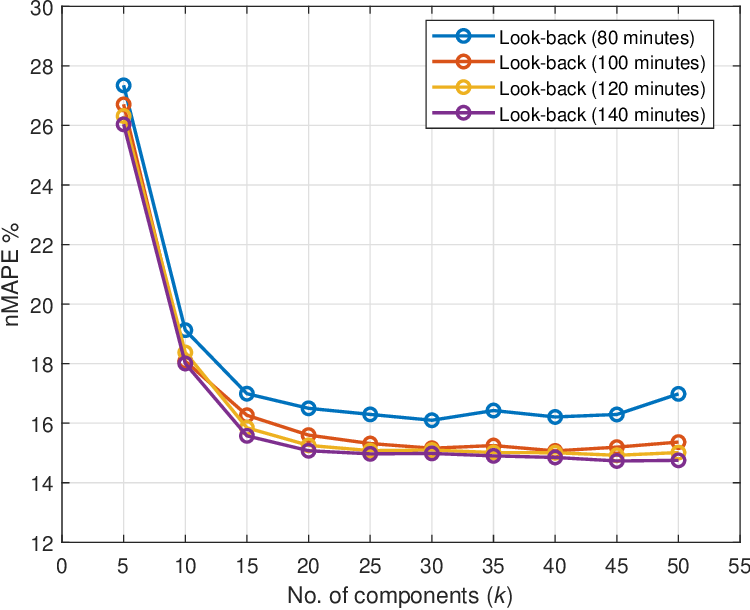}\hfill

}
\subfigure[~ASI-16]{
\includegraphics[scale=0.6]{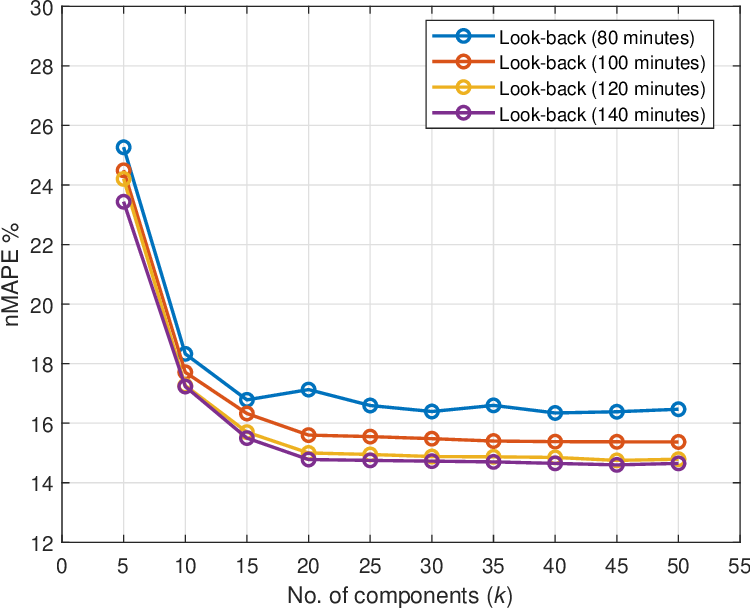}\hfill

}

\caption{Normalized mean absolute percentage error (nMAPE) vs. \(k\) value for different look-back~intervals.}
\label{LSA}
\end{figure}

As mentioned in Section~\ref{sec3}, both LSA and sample look-back are considered for forecasting. Figure~\ref{LSA} show the tuning process for both \(k\) and look-back intervals when the two datasets, TSI-880 and ASI-16, are considered. Tuning the sample look-back interval gives rise to a trade off between the look-back time and the accuracy. For~example, in~the case of a 12-sample look-back, and~with 10 min between each frame, 2 h will be the prediction latency; in other words, the~model will wait 2 h to produce the first forecast prediction (1--4 h ahead). On~the other hand, increasing the look-back period will increase the accuracy but up to a certain limit. Therefore, due to the importance of forecasting GHI as early as possible for the application at hand, Figure~\ref{LSA} suggests that the  optimal parameters are \(k\) = 20 with look-back of 120~min.

\begin{figure}[H]
\centering
\captionsetup{justification=centering}

\includegraphics[scale=0.35]{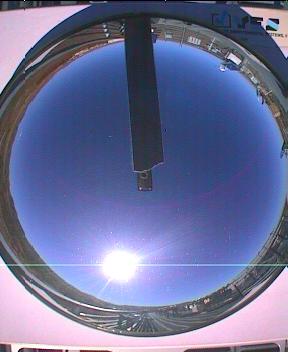}
\includegraphics[scale=0.35]{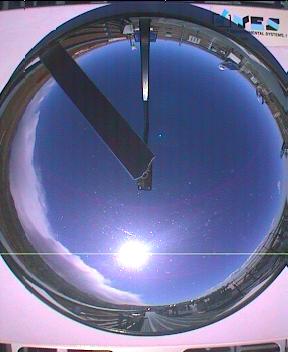}

\caption{ Images from 2015 and 2016, respectively, where the sun-tracker has stopped~working.}
\label{mistrack}
\end{figure}

The nowcasting and forecasting results for the two datasets are reported in Tables~\ref{nMAPE} and \ref{RMSE}. For~the~TSI-880 dataset in years 2015 and 2016, the~results of the proposed approach are demonstrated along with the original VGG16 deep learning framework~\cite{simonyan2014very}, as~well as the approach of~\cite{siddiqui2019deep}. In~nowcasting, the~approach of~\cite{siddiqui2019deep} augmented the training of their model with auxiliary weather parameters (average wind speed, relative humidity, barometric pressure, air temperature, sun position (\(z\)), and~clear sky prediction). We observe that our model achieved comparable results for nowcasting with respect to the state-of-the-art models, as~shown in Table~\ref{nMAPE}. Additionally, applying the proposed prediction algorithms to the first 10 years (2004 to 2014), excluding both the years 2015 and 2016, will produce superior results. The~reason is due to the following: during the period from May 2015 to December 31, 2016, the~sun tracker has stopped working, as~shown in Figure~\ref{mistrack}. Therefore, the~captured images are greatly affected by the sun. Contrary, the~approach of~\cite{siddiqui2019deep} uses more robust techniques of covering a higher receptive region of sky images with cloud movement to extract relevant features from an image, therefore mitigating the effect of the sun tracker. This difference is apparent in forecasting since the proposed approach uses sample look-back for prediction. As~shown in Table~\ref{nMAPE}, the~performance of the proposed prediction algorithms did not perform well in comparison with the approach of~\cite{siddiqui2019deep} when the sun tracker is inactive. On~the other hand, when 2 years of data are randomly selected for testing, the~performance of the proposed approaches is improved by a significant~difference.

\unskip

\begin{table}[H]
\centering
\caption{nMAPE for nowcasting and forecasting results using different~methods. }

\scalebox{.9}[0.9]{\begin{tabular}{@{}cccccccccc@{}}

\toprule

\multirow{2}{*}{\textbf{Dataset}} & \multirow{2}{*}{\textbf{Method}} & \multirow{2}{*}{\textbf{Test Period}} && \multirow{2}{*}{\textbf{Nowcasting nMAPE (\%) }}  & \multicolumn{4}{c}{\textbf{Forecasting nMAPE (\%)}}\\\cmidrule{6-9} &&&&& \textbf{+1 hr} & \textbf{+2 hr} & \textbf{+3 hr} & \textbf{+4 hr}\\  \cmidrule{1-9}

\multirow{10}{*}{TSI-880} & \multirow{2}{*}{VGG16~\cite{simonyan2014very}}
&  2015 && 21.0 & -  & - & - & -  \\
&&  2016 && 21.9 & -  & -  & - & -   \\\cmidrule{2-9}
& \multirow{2}{*}{A. Siddiqui et al \cite{siddiqui2019deep}}
&  2015 && 14.6 & 17.9   & 25.2 & 31.6 & 39.1  \\
&&  2016 && 15.7 & 16.9   & 25.0  & 31.9 & 39.5   \\\cmidrule{2-9}
&\multirow{3}{*}{KNN}
& 2015 && 17.51 & 36   & 38.9 & 41.5 & 44.4\\
&&  2016 &&16.79 & 36.5   & 39.5  & 42.1 & 45.2\\
&&  2 years (random) &&10.2 & 14.9   & 16.7 & 18.7 & 21.1\\ \cmidrule{2-9}
&\multirow{3}{*}{RF}
&2015 &&14.1 & 30.8   & 34.2 & 36.9 & 40.1\\
&&  2016 &&14.8&31.4   & 34.7  & 37.5 & 40.6\\
&&  2 years (random) && 9.8& 21.9  & 24.9 & 27.8 & 30.6 \\\midrule
\multirow{2}{*}{ASI-16} & KNN & 1 year (random) &&14.5 &14.7 &15.8 & 16.6 & 18.4 \\\cmidrule{2-9}
& RF & 1 year (random) && 13.35 &23.5 &25.5 &27.6 & 30.5 \\

\bottomrule

\label{nMAPE}
\end{tabular}}
\end{table}
\unskip

\begin{table}[H]
\centering
\caption{Root mean square error (RMSE) and normalized root mean square error (nRMSE) for nowcasting and forecasting results using the proposed~methods. }

\scalebox{.9}[0.9]{\begin{tabular}{@{}ccccccccc@{}}

\toprule

\multirow{2}{*}{\textbf{Dataset}} & \multirow{2}{*}{\textbf{Method}} & \multirow{2}{*}{\textbf{Test Period}} & \multirow{2}{*}{\textbf{Performance Metric}} & \multirow{2}{*}{\textbf{Nowcasting }}  & \multicolumn{4}{c}{\textbf{Forecasting}}\\\cmidrule{6-9} &&&&& \textbf{+1 hr} & \textbf{+2 hr} & \textbf{+3 hr} & \textbf{+4 hr}\\
\midrule

\multirow{4}{*}{TSI-880}
&\multirow{2}{*}{KNN}

& \multirow{2}{*}{2 years (random)}  & RMSE (W/m\textsuperscript{2})  &
71.0 & 122.2  & 137.4 & 151.1 & 164.4\\
&&& nRMSE (\%)  &4.4 & 7.7  & 9.6 & 11.2 & 12 \\\cmidrule{2-9}
&\multirow{2}{*}{RF}

& \multirow{2}{*}{2 years (random)}  & RMSE (W/m\textsuperscript{2})  &
64.7 & 141.8  & 158.9 & 171.2 & 183.2\\
&&& nRMSE (\%)  & 4& 8.9 & 11.1 & 12.7 & 13.5 \\
\midrule

\multirow{4}{*}{ASI-16}                         &\multirow{2}{*}{KNN}

& \multirow{2}{*}{1 years (random)}  & RMSE (W/m\textsuperscript{2})  &
112.3 & 116.7  & 127.6 & 132.3 & 143.8\\
&&& nRMSE (\%)  & 8.5 & 8.9  & 9.3 & 10.2 & 11.3 \\\cmidrule{2-9}
&\multirow{2}{*}{RF}

& \multirow{2}{*}{1 years (random)}  & RMSE (W/m\textsuperscript{2})  &
111.4 & 141.3  & 156.3 & 164.6 & 173.3\\
&&& nRMSE (\%)  & 8.1 & 10.8  & 11.4 & 12.7 & 13.7 \\

\bottomrule

\label{RMSE}
\end{tabular}}
\end{table}

\begin{figure}[H]
\captionsetup{justification=centering}
\centering
\includegraphics[scale=0.75]{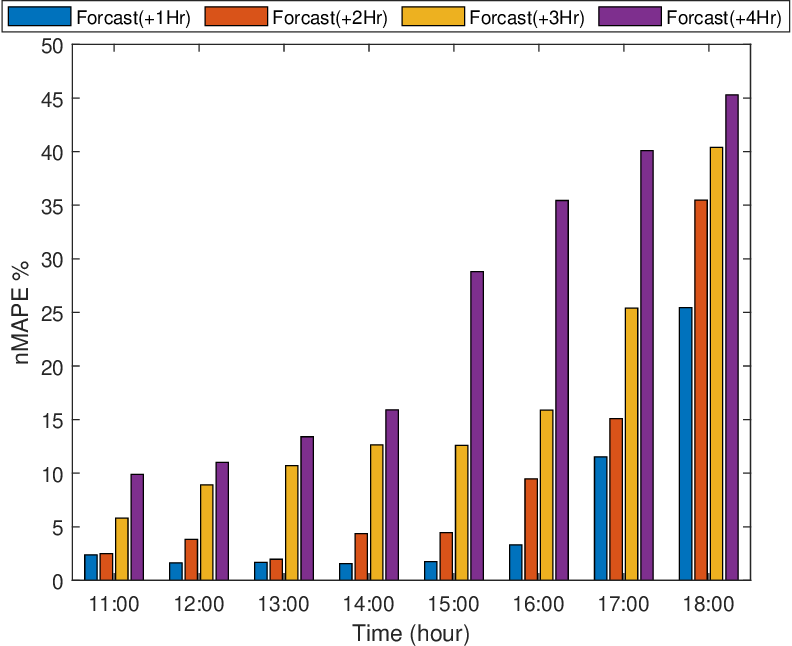}

\caption{Hourly nMAPE forecast of +1, +2, +3,
and +4 h. }
\label{Forecast}

\end{figure}
\begin{figure}[H]
\centering

\subfigure[~RMSE]{
\includegraphics[scale=0.595]{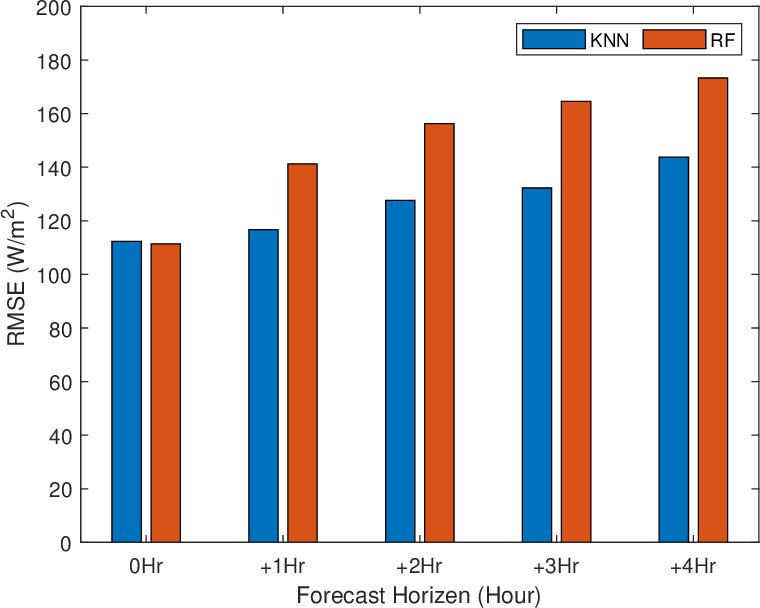}

}
\subfigure[~nRMSE]{
\includegraphics[scale=0.595]{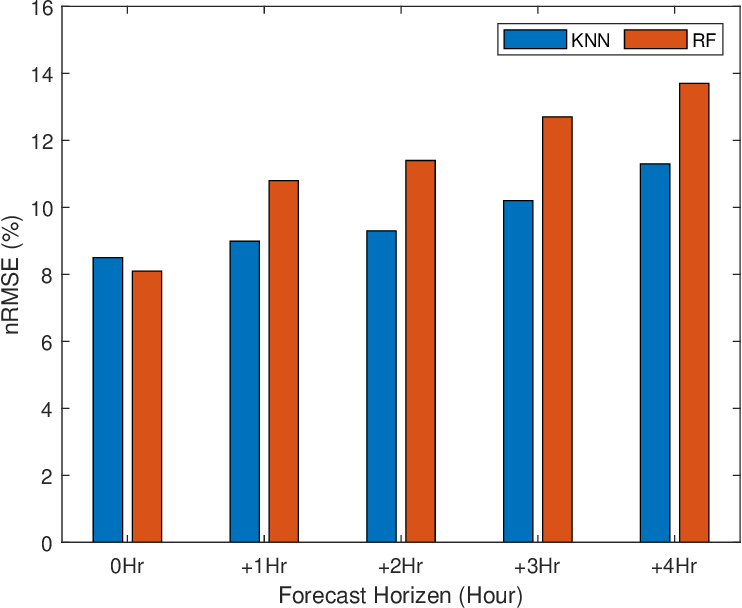}

}

\caption{RMSE and nRMSE prediction errors for K-nearest neighbors (KNN) and~ Random Forest~(RF). }
\label{nRMSE}
\end{figure}

With reference to Table~\ref{nMAPE}, for~the TSI-880 dataset and KNN model, the~nMAPE values are 14.9\%, 16.7\%, 18.7\%, and~21.1\% for 1--4 h ahead forecasts, respectively, while, for the ASI-16 dataset, the~nMAPE values are 14.7\%, 15.8\%, 16.6\%, and~18.4\%, respectively. The~results of the KNN model for the two datasets are very close, which further confirms the effectiveness of proposed prediction approach. A~second note is that the RF algorithm performs better in nowcasting, while the KNN algorithm is the best in forecasting.
Figure~\ref{Forecast} shows the ahead-of-time forecasting errors in an hourly fashion for the KNN model. The~error increases for larger forecast horizons as well as for later hours in the day. Table~\ref{RMSE} reports the prediction accuracy using RMSE and nRMSE. Figure~\ref{nRMSE} shows the RMSE and nRMSE for both KNN and RF. Note that, for the TSI-880 dataset and KNN model, the~nRMSE values are 7.7\%, 9.6\%, 11.2\%, and~12\% for 1--4 h ahead forecasts, respectively, while, for the ASI-16 dataset, the~nRMSE values are 8.9\%, 9.3\%, 10.2\%, and~11.3\%, respectively.


\unskip

\begin{figure}[H]
\centering

\subfigure[~Sunny]{\hspace{-3.5mm}
\includegraphics[scale=0.5]{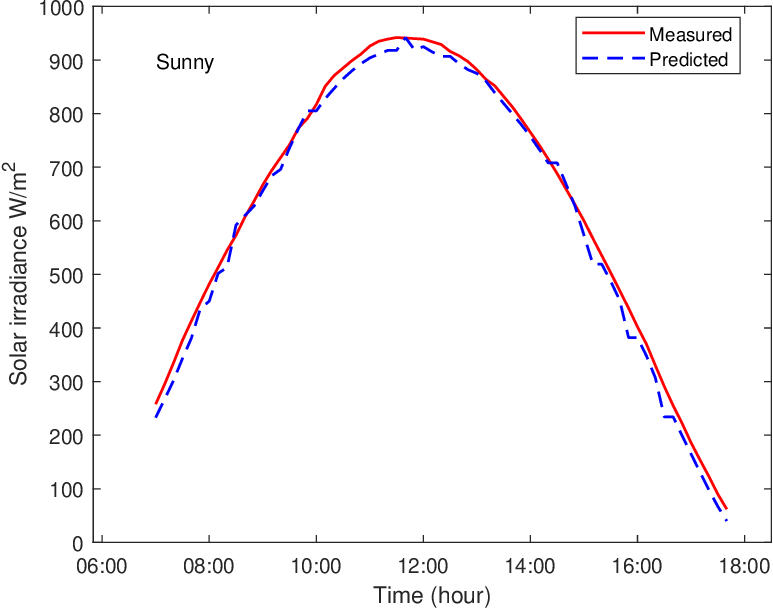}

}
\subfigure[~Cloudy]{
\includegraphics[scale=0.5]{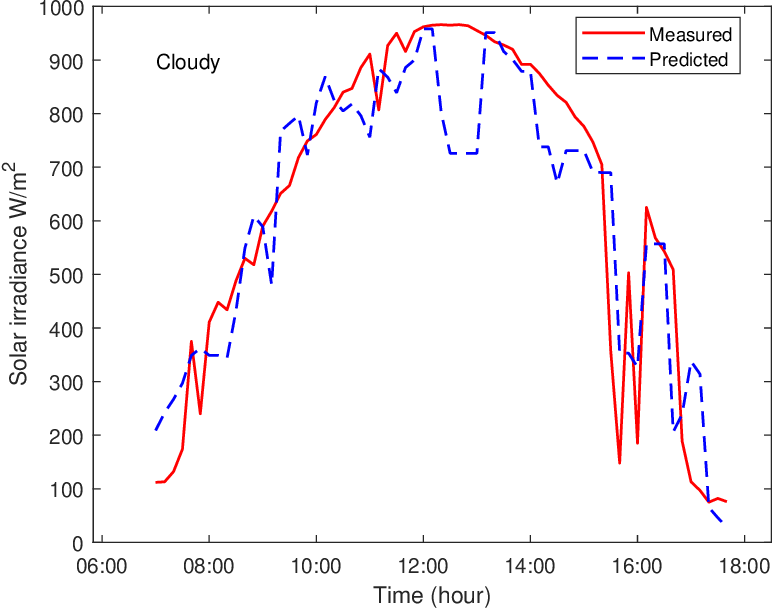}

}
\subfigure[~Rainy]{\hspace{5mm}
\includegraphics[scale=0.5]{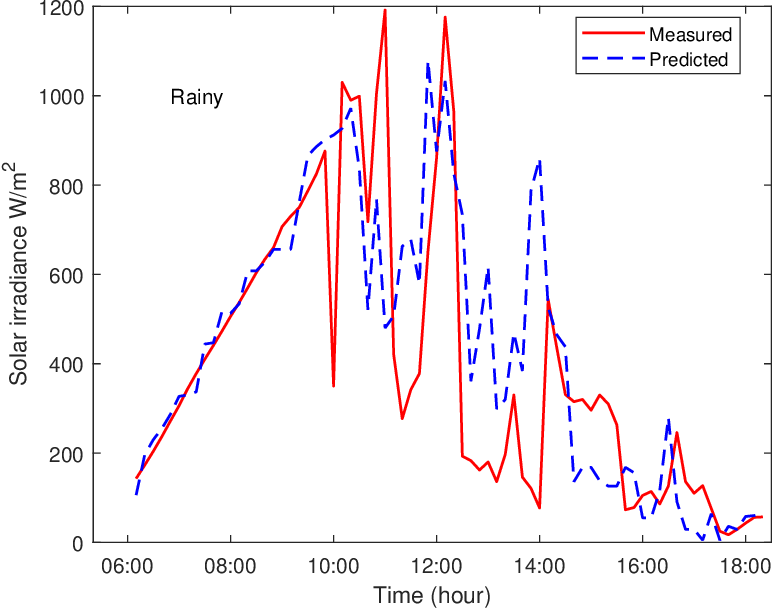}

}
\subfigure[~Sunny and cloudy]{
\includegraphics[scale=0.5]{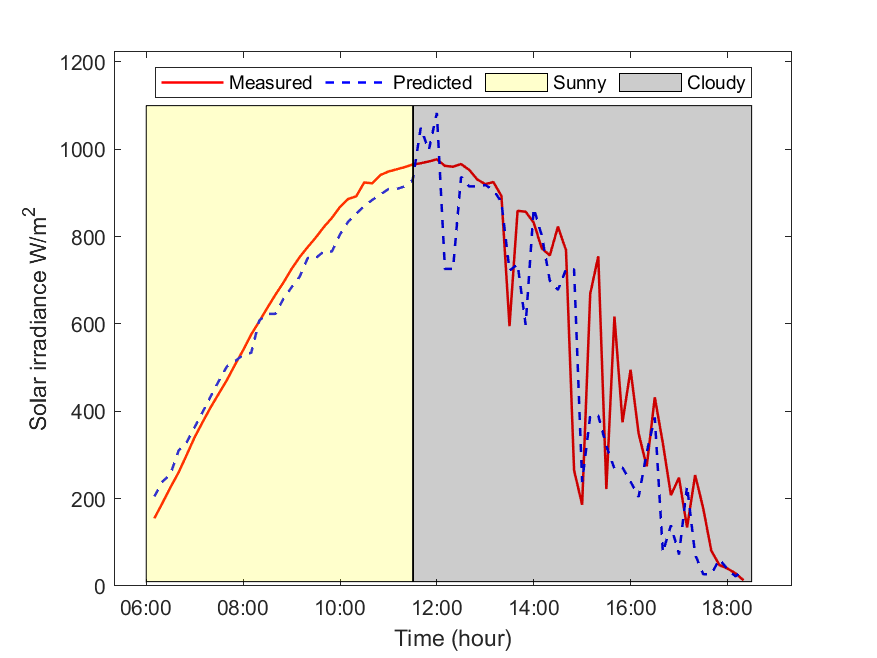}

}
\caption{Measured vs. forecasted hourly solar irradiance for three types of weather~conditions. }
\label{weather}
\end{figure}

Three types of weather conditions are considered to compare the predicted solar irradiance of the proposed models with the measured values. Figure~\ref{weather}a  shows close agreement between the predicted and measured GHI values for a sunny day. Figure~\ref{weather}b,c show the predicted and measured values for cloudy and rainy days, with~a noticeable discrepancy, which is more pronounced between the predicted and measured values for the rainy day. This, in part, is due to the rapid changes in the hourly irradiance values during the day. The~nMAPE values for sunny, cloudy and rainy days are 3.1\%, 14.3\%, and~20.5\%, respectively. Furthermore, Figure~\ref{weather}d illustrates the effect of the rapid change in hourly irradiance on the prediction as the weather shifts from sunny to~cloudy.


\par It is relevant to mention here that the approach of~\cite{siddiqui2019deep} is computationally more expensive when compared to our proposed prediction algorithms. In~particular,  the~proposed architecture of~\cite{siddiqui2019deep} uses a sky image of dimensions (64, 64, 3) as an input to a CNN-based model. This model is obtained by performing the ablation of layers from the original VGG16 architecture~\cite{simonyan2014very}, which, compared to our model, is a very deep and computationally complex approach. This CNN stage, aided by auxiliary data, is only used to predict a single unit of solar irradiance (nowcasting). For~forecasting, a~two-tier-long short-term memory (LSTM) neural network has been considered to utilize the model of the CNN stage to obtain historical full-sky representations (look-back) and produce ahead-of-time forecasts.  The~computational  complexity of CNN convolution layers can be approximated by~\cite{he2015convolutional}:
\begin{equation}
O(\sum_{l=1}^{d}n_{l-1}\times s_{l}^{2}\times n_{l}\times m_{l}^{2} )
\end{equation}
where \(d\) is the number of convolution layers, \(l\) is the index of a convolution layer, \(n_{l}\) is the number of filters, \(n_{l-1}\) is the number of input channels of \(l\)th layer, \(s_{l}\) is the spatial size of the filter, and~\(m_{l}\) is the spatial size of the output feature map. In~\cite{siddiqui2019deep}, CNN and LSTM were used to obtain ahead-of-time forecasting. The~time complexity for the LSTM stage is considered to be $O(1)$ \cite{tsironi2017analysis}. The~pooling and fully connected layers (FCL) take about 10\% of the computational time~\cite{he2015convolutional}. On~the other hand, the~computational complexities of RF and KNN regressors are $O( n_{trees}\times \log(N) )$ and $O(N\times p )$, respectively, where \(N\) is the number of training samples, \(p\) is the number of features, and~\(n_{trees}\) is the number of trees~\cite{deng2016efficient,louppe2014understanding}. Using the values of relevant parameters of the predictors under consideration in Section~\ref{sec3.2}, we find that the computational complexities of KNN and RF are reduced by 30\% and 95\% compared to those of the CNN-based approach, respectively.

\section{Conclusions and Future~Work}
\label{sec5}
Accurate solar irradiance forecasting is crucial for the stability of the power grid.
In this paper, a~solar irradiance forecast approach was
proposed, which combines ML methods with
dimensionality reduction techniques. The~learning algorithm
is able to perform forecasting for solar irradiance up to 4 h ahead.
Two different datasets were used in this study to comprehensively evaluate
the performance of proposed prediction approaches. In~addition, three statistical metrics were used to assess the performance of the proposed approaches. It has been found that the proposed KNN-based approach can achieve the following performance. For~TSI-880 and ASI-16 hourly forecasts, the~nMAPE is 14.9\% and 14.7\%, the~RMSE is 122.2 W/m\textsuperscript{2} and 116.7 W/m\textsuperscript{2}, and~the nRMSE is 7.7\% and 8.9\%, respectively. These results achieved   are  competitive  compared to the state-of-the-art algorithms, while utilizing computationally efficient techniques for the nowcasting and forecasting of surface irradiance. In~particular, the~proposed KNN-based approach achieves better computational complexity, which is reduced by 30\% of that of the state-of-the-art algorithms.
\par \textcolor{black}{A possible direction for future work is to implement the proposed prediction algorithms on a low-cost sky imaging system. An~initial investigation indicates that a Raspberry Pi single-board computer connected to a programmable, high-resolution Pi camera (with a fisheye lens for a wide field of view) can be used. The~Raspberry Pi is a powerful platform because of its processing capabilities, its lean design, and~low power requirement. For~performance evaluation purposes, a~precision pyranometer is required to provide the ground truth values of GHI. The~pyranometer is a device used to measure the irradiance (W/m\textsuperscript{2}) on a plane surface, which results from the direct solar radiation and the diffuse radiation incident from the hemisphere above. To~accurately assess an all-sky imaging system, sufficient data acquisition needs to be performed to cover different weather conditions (sunny, cloudy, rainy, and~dusty).}

\vspace{6pt}



\authorcontributions{Conceptualization, A.A.-l., O.T. and K.E.; methodology, A.A.-l., O.T. and K.E.; software, A.A.-l., O.T. and K.E.; validation, A.A.-l. and O.T.; formal analysis, A.A.-l., O.T. and K.E.; investigation, A.A.-l., O.T. and K.E.; resources, A.A.-l., O.T. and K.E.; data curation, A.A.-l.; writing---original draft preparation, A.A.-l., O.T. and K.E.; writing---review and editing, A.A.-l., T.A.A. and S.A.A.; visualization, A.A.-l., O.T. and K.E.; supervision, T.A.A. and S.A.A.; project administration, T.A.A. and S.A.A.; funding acquisition, S.A.A. All authors have read and agreed to the published version of the~manuscript.}

\funding{This work was supported by the Researchers Supporting Project number (RSP-2020/46), King Saud University, Riyadh, Saudi Arabia.}

\acknowledgments{The authors would like to acknowledge the Researchers Supporting Project at King Saud University.}

\conflictsofinterest{The authors declare no conflicts of interest. The~funders had no role in the design of the study; in the collection, analyses, or~interpretation of data; in the writing of the manuscript, or~in the decision to publish the~results.}







\reftitle{References}

\end{document}